\documentclass[12pt,a4paper]{article} 
\usepackage{graphics}

\begin{document}

\begin{center}
{\Large Improvements of Real Coded Genetic Algorithms}\\
\vspace{1mm}
{\Large Based on Differential Operators}\\
\vspace{1mm}
{\Large Preventing the Premature Convergence }\\
\vspace{6mm}

 O.\,Hrstka$^1$ and A.\,Ku\v{c}erov\'a$^2$
\\
\vspace*{6mm} $^1$
Computing and Information Center, Faculty of Civil 
Engineering, Czech Technical University in Prague, 
Th{\' a}kurova 7, Prague, Czech Republic
\\
\vspace{1mm}
$^2$
Department of Structural Mechanics, Faculty of Civil 
Engineering, Czech Technical University in Prague,
Th{\' a}kurova 7, Prague, Czech Republic
\end{center}
\vspace*{5mm}
 
\begin{abstract}
This paper presents several types of evolutionary algorithms~(EAs)
used for global optimization on real domains. The interest has been
focused on multimodal problems, where the difficulties of a premature
convergence usually occurs. First the standard genetic algorithm (SGA)
using binary encoding of real values and its unsatisfactory behavior
with multimodal problems is briefly reviewed together with some
improvements of fighting premature convergence. Two types of real
encoded methods based on differential operators are examined in
detail: the differential evolution~(DE), a very modern and effective
method firstly published by R.\,Storn and K.\,Price~\cite{Storn}, and
the simplified real-coded differential genetic algorithm SADE proposed
by the authors~\cite{Sade}. In addition, an improvement of the SADE
method, called CERAF technology, enabling the population of solutions
to escape from local extremes, is examined. All methods are tested on
an identical set of objective functions and a systematic comparison
based on a reliable methodology~\cite{Francouzi} is presented. It is
confirmed that real coded methods generally exhibit better behavior on
real domains than the binary algorithms, even when extended by several
improvements. Furthermore, the positive influence of the differential
operators due to their possibility of self-adaptation is
demonstrated. From the reliability point of view, it seems that the
real encoded differential algorithm, improved by the technology
described in this paper, is a universal and reliable method capable of
solving all proposed test problems.
\end{abstract}

\section{Introduction}

At present, genetic algorithms belong to the most modern and most
popular optimization methods. They follow an analogy of processes that
occur in living nature within the evolution of live organisms during a
period of many millions of years. The principles of genetic algorithms
were firstly proposed by J.\,H.\,Holland~\cite{Holland}; the books of
D.\,E.\,Goldberg \cite{Goldberg} and Z.\,Michalewicz
\cite{Michalewicz} are the most popular publications that deal with this
topic. Genetic algorithms have been successfully used to solve
optimization problems in combinatorics (see \cite{Grefenstette}) as
well as in different engineering tasks, see for example
\cite{Matej1,Matej2,Rafiq}. \par

Unlike the traditional gradient optimization methods, genetic
algorithms operate on a set of possible solutions~(``chromozomes''),
called ``population''. In the basic scheme, chromozomes are
represented as binary strings. This kind of representation seems to be
very convenient for optimization problems in combinatoric area (e.g.,
the traveling salesman problem). Nevertheless, we usually deal with
real valued parameters in engineering and scientific problems.  The
mapping of real values onto binary strings usually used within
standard genetic algorithms may cause serious difficulties.  As a
result, this concept of optimization leads to an unsatisfactory
behavior, characterized by a slow convergence and an insufficient
precision, even in cases where the precision is especially in
focus. Of course, the development of genetic algorithms has brought
several proposals to solve these difficulties to optimize problems on
real domains using binary algorithms. \par

Another possibility is to develop a genetic algorithm (or other
evolutionary algorithm) that operates directly on real
values~\cite{Michalewicz}. In this case, the biggest problem is how to
propose genetic operators. One of them is to use so-called
differential operators that are based on determining mutual distances
of chromozomes -- which are real vectors instead of binary strings in
this approach. \par

This paper studies two evolutionary optimization methods based on
differential operators with reference to the standard and improved
genetic algorithms using binary encoding. In particular, the
differential evolution proposed by
R.\,Storn~and~K.\,Price~\cite{Storn,Storn-WWW} and the simplified real
coded differential genetic algorithm~\cite{Sade,Sade-WWW} are examined
in very details. 

Although the outstanding ability of genetic algorithms to find global
optima of multimodal functions (functions which have several local
extremes) is usually cited in the GA literature, it seems that both the binary genetic
algorithms and the real coded ones tend to premature converge and to
fall into local extremes, mainly in high dimensional cases. To fight
this difficulty, we have proposed so-called CERAF method. \par 

As the reference, test results for binary encoded algorithms from the
outstanding article of J.\,Andre, P.\,Siarry and T.\,Dognon
\cite{Francouzi} were selected. These results come from two variants
of binary GAs: the standard genetic algorithm and the version extended
by several improvements that were documented in the same
publication. The set of twenty test functions was used to classify
reliability and performance of individual methods. In particular, the
reliability is defined as a probability of finding the global extreme
of a multimodal function while the performance is measured by the
convergence rate of an optimization method. Since this methodology is
able to filter out the influence of random circumstances we have used
the same criteria to quantify the robustness and the efficiency of
real encoded optimization methods.

\section{Binary coded genetic algorithm}
%
Although the present paper deals mainly with real encoded evolutionary
algorithms, we present a brief description of binary genetic
algorithms, their limitations and possible improvements for the sake
of further reference. \par

\subsection{Binary encoding}
%
A binary genetic algorithms can be simply characterized by the binary
encoding of possible solutions and appropriate binary genetic
operators. The traditional binary genetic algorithms represent
possible solutions as binary strings, usually derived from a division
of the investigated interval into a several sub-intervals with a
specified, usually rather limited, precision. The fact that different
bits in the binary string have different importance depending on their
position in the string is the serious problem of this type of
encoding. This disadvantage can be resolved by several improvements,
see Section~\ref{ssec:isbga} for particular examples and references to
the literature.

\subsection{Scheme of genetic algorithm and genetic operators}
%
As the first step it is necessary to generate (in most cases randomly)
the starting population of possible solutions that are assigned the
values of the optimized (or so-called fitness) function. Then, the
sequential loop is repeated until a stopping criterion is reached:

\begin{enumerate}
\item Create a prescribed number of new individuals (chromozomes)
using genetic operators of crossing-over and mutation.

\item Values of fitness function are assigned to new individuals. 

\item The population size is decreased to the original value using 
selection operator. 
\end{enumerate}

\begin{figure}[h]
\begin{center}
\scalebox{.5}{
\includegraphics{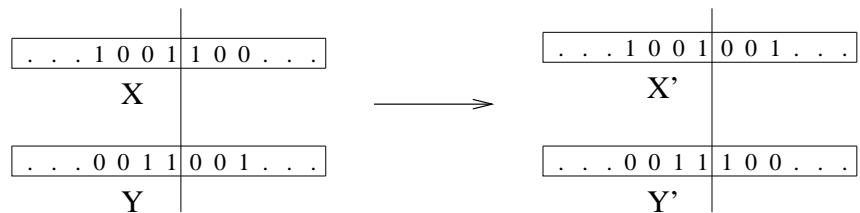}}
\end{center}
\caption{Crossing-over operator.}
\label{crossover}
\end{figure}

In the following, we present a sketchy description of basic genetic
operators:

\begin{description}
\item [mutation] -- the principle of this operator is an alteration of
one or more bits in the binary string \cite{Foo}; a parameter which
gives a probability of performing this operation with a certain
chromozome, is introduced,

\item [crossing-over] -- this operator chooses two chromozomes,
so-called parents, and then creates their two descendants (children)
using the following operation: it selects a position inside the binary
string and starting from this position exchanges the remaining parts
of the two chromozomes (see Figure~\ref{crossover}). The individuals
subject to crossing-over are selected by an appropriate sampling
method, which is not neccesarilly identical to the selection method
employed in the next step.

\item [selection] -- this operation selects the individuals that
should ``survive'' into the next generation from the whole
population. See, e.g.,~\cite{Baker,Goldberg} for a comprehensive list
of different variants of selection schemes.

\end{description}  

\subsection{Improvements of the standard binary genetic algorithm}\label{ssec:isbga}
%
A lot of improvements of the standard binary genetic algorithm that
aim at suppressing the premature convergence have been proposed by
different authors, starting from different encoding, e.g., the
well-known {\em Gray code}~\cite{Grayscale}; proceeding with {\em
threshold genetic algorithm} with varying mutation
probability~\cite{Cinani}. Another possibilities are various
adaptations of the crossing-over and the mutation or an introduction
of local, gradient-based operators such as {\em gradient
optimizer\/}~\cite{Davis} or {\em evolutionary gradient operator\/}
proposed in~\cite{Yamamoto}. In particular, in the reference
study~\cite{Francouzi} the authors performed the testing computations
of the standard genetic algorithm extended by {\em adaptive rescaling}
of the investigated area and by introducing the {\em
scale-factor\/}. For the sake of clarity, a more detailed description
of these improvements follows:

\begin{itemize}

\item the {\em adaptive rescaling} of investigated domain: the area
where the method searches for the optimum is diminished into the
regions around several best chromozomes; it makes possible to reach
very good precision even if the division of investigated interval is
rather rough because with the decreasing searched range the division
becomes more and more refined,

\item introducing so-called {\em scale-factor} (SF) which influences
selection of individuals to be subject to the crossing-over; at the
beginning the worse individuals gain higher probability and the better
ones gain lower probability contrary to the standard genetic
algorithm; with successive generations this parameters decreases and
for the last generation the selection works in the same manner as for
the standard version.

\end{itemize} 

\subsection{Genetic algorithm testing methodology}
%
The methodology proposed in \cite{Francouzi} minimizes an influence of
random circumstances and different power of the used computers. In
particular, the computation is run $100$ times for each function of
the test set. The number of successful runs is then taken as the
probability of the success (the computation is considered to be
successful if the difference between the best value found by the
algorithm and the theoretical optimum is less than $1\%$ of the
optimum value, or a distance is less than $0.1$ if the theoretical
optimum is zero). If $500$ generations pass and the optimum is still
not reached, the computation is treated as a failure. For the cases
where the amount of successful runs is greater than zero, the average
fitness call number is also given. For the sake of completeness, we
list a (corrected) set of twenty test functions in
Appendix~\ref{sec:appa} while the results of the binary genetic
algorithm testing are shown in Table~\ref{results}. \par

\begin{table}[h]
\begin{center}
\begin{tabular}{lrrrrrrrrrrr}
\hline
& &
\multicolumn{2}{c}{{\bf SBGA}} & 
\multicolumn{2}{c}{{\bf EBGA}} &
\multicolumn{2}{c}{{\bf DE}} &
\multicolumn{2}{c}{{\bf SADE}} \\
\bf Test function & $N$ & 
\bf SR \% & \bf NFC & 
\bf SR \% & \bf NFC & 
\bf SR \% & \bf NFC &
\bf SR \% & \bf NFC \\
\hline
F1 & 1 & 100 & 5566 & 100 & 784 & 100 & 52 & 100 & 72 \\
F3 & 1 & 100 & 5347 & 100 & 744 & 100 & 98 & 100 & 88 \\
Branin & 2 & 81 & 8125 & 100 & 2040 & 100 & 506 & 100 & 478 \\
Camelback & 2 & 98 & 1316 & 100 & 1316 & 100 & 244 & 100 & 273 \\
Goldprice & 2 & 59 & 8125 & 100 & 4632 & 100 & 350 & 100 & 452 \\
PShubert1 & 2 & 63 & 7192 & 100 & 8853 &  83 & 1342 & 100 & 2738 \\
PShubert2 & 2 & 59 & 7303 & 100 & 4116 &  90 & 908  & 100 & 1033 \\
Quartic & 2 & 83 & 8181 & 100 & 3168   &  97 & 313  & 100 & 425 \\
Shubert & 2 & 93 & 6976 & 100 & 2364   &  94 & 10098 & 100 & 585 \\
Hartman1 & 3 & 94 & 1993 & 100 & 1680  & 100 & 284 & 100 & 464 \\
Shekel1 & 4 & 1 & 7495 & 97 & 36388    & 72 & 1968 & 99 & 61243 \\
Shekel2 & 4 & 0 & - & 98 & 36774       & 91 & 1851 & 100 & 17078 \\
Shekel3 & 4 & 0 & - & 100 & 36772      & 89 & 1752 & 99 & 11960 \\
Hartman2 & 6 & 23 & 19452 & 92 & 53792 & 16 & 4241 & 67 & 2297 \\
Hosc45 & 10 & 0 & - & 2 & 126139 & 100 & 1174 & 100 & 6438 \\
Brown1 & 20 & 0 & - & 0 & - & 100 & 65346 & 95 & 163919 \\
Brown3 & 20 & 5 & 8410 & 5 & 106859 & 100 & 41760 & 100 & 43426 \\
F5n & 20 & 0 & - & 100 & 99945 & 96 & 38045 & 66 & 17785 \\
F10n & 20 & 0 & - & 49 & 113929 & 90 & 71631 & 47 & 110593 \\
F15n & 20 & 0 & - & 100 & 102413 & 100 & 44248 & 93 & 28223  \\
\hline
\end{tabular}
\end{center}
\caption{Comparison of results of investigated methods. SR = success
rate, NFC = average number of function calls, $N$ = dimension of the
problem, SBGA = Standard Binary GA, EBGA = Extended Binary GA, DE =
Differential Evolution, SADE = Simplified Atavistic DE}
\label{results}
\end{table}

\section{Differential evolution}
%
This section opens the main topic of the present work -- a search for
improvements of real coded genetic algorithms aimed at resolving the
premature convergence. To this end, a thorough description of the
differential evolution, which is the stepping stone of our
improvements, is presented. \par

The differential evolution belongs to the wide group of evolutionary
algorithms. It was invented as the solution method for the Chebychev
trial polynomial problem by R.\,Storn and K.\,Price~\cite{Storn}. It
is a very modern and efficient optimization method essentially relying
on so-called differential operator, which works with real numbers in
natural manner and fulfills the same purpose as the crossing-over
operator in the standard genetic algorithm. \par

\subsection{The differential operator}
%
The typical differential operator has the sequential character: Let
$CH_i(t)$ be the {\em i}-th chromozome of a generation {\em t}

\begin{equation}
CH_i(t)=(ch_{i1}(t),ch_{i2}(t),\ldots,ch_{in}(t)), 
\end{equation}

where $n$ is the chromozome length (which equals to the number of
variables of the fitness function in the real encoded case). Next, let
$\Lambda$ be a subset of $\{1,2,\ldots,n\}$\footnote{It may be chosen
randomly, for example.}. Then, for each $j\in\Lambda$,
\begin{eqnarray}
ch_{ij}(t+1)=ch_{ij}(t) & + & F_1\left( ch_{pj}(t)-ch_{qj}(t)\right) 
\nonumber \\ 
& + & F_2\left( ch_{\mathrm{best}j}(t)-ch_{ij}(t)\right), 
\end{eqnarray}
and for each $j\notin\Lambda$ 
\begin{equation}
ch_{ij}(t+1)=ch_{ij}(t), 
\end{equation}
where $ch_{pj}$ and $ch_{qj}$ are the {\em j}-th coordinates of two
randomly chosen chromozomes and $ch_{\mathrm{best}j}$ is the {\em
j}-th coordinate of the best chromozome in generation~{\em t}. $F_1$
and $F_2$ are random coefficients usually taken from interval
$(0,1)$. \par

\subsection{The differential evolution algorithmic scheme}
%
The differential evolution can be understood as a stand-alone 
evolutionary method or it can be taken as a special case of the 
genetic algorithm. The algorithmic scheme is similar to the genetic 
algorithms but  it is much simpler:
\begin{enumerate}

\item At the beginning the initial population is created (e.g.,
randomly) and the fitness function value is assigned to each
individual.  

\item For each chromozome in the population, its possible replacement 
is created using the differential operator discussed above.

\item Each chromozome in the population has to be compared with its
possible replacement and if an improvement occurs, it is replaced.

\item Steps 2 and 3 are repeated until a stopping criterion is 
reached.

\end{enumerate}

As could be seen, there are certain different features in contrary to 
the standard genetic algorithm, namely:
\begin{itemize}
\item the crossing-over is performed by applying the differential
operator,

\item selection for crossing-over, e.g. by the roulette wheel
method, is not performed; the individuals subjected to the
differential operator are chosen purely randomly,

\item selection of the individuals to survive is simplified: each
chromozome has its possible replacement and only the worse in terms
of fitness is replaced,

\item the mutation operator is omitted.

\end{itemize}

\subsection{Test computations}
%
The differential evolution was examined on previously introduced set
of test functions as the binary encoded algorithms and using the same
methodology. All computations were performed with identical parameters
setting:
\begin{equation}
F_1=F_2=0.85 \quad\mathrm{and}\quad \Lambda=\{ 1,2,\ldots,n \}.    
\end{equation}
The population size was set to $pop=10n$ for all examined
functions. The results are presented in Table~\ref{results}. \par

Comparing the results of the differential evolution to those reported
in~\cite{Francouzi}, several interesting findings are apparent.
First, the differential evolution shows substantially better
reliability solving the most difficult functions where even the
extended binary genetic algorithm failed. Only for the {\em Hartman
2\/} function the result is unsatisfactory (only $16\%$). In all other
cases the probability of success is better than $70\%$ and in $10$
cases it reaches $100\%$ (including {\em Hosc~45\/} and both {\em
Brown 1\/} and {\em 3\/} functions, for which the extended binary
genetic algorithm failed). On the other hand, for several functions,
where the binary algorithm shows $100\%$~success, the differential
evolution rests at about $70$-$95\%$. \par

Another effect that is evident from the comparison is the fact that
the differential evolution is able to find a solution with the same
precision much faster (for example, 52 fitness calls contrary to 784
fitness calls using the extended binary algorithm for the {\em F1\/}
function). We suppose that this improvement is a consequence of the
very good precision adaptability of the differential operators as all
alterations of the chromozomes are determined from their mutual
distances, see Figure~\ref{operator}. \par

\section{Differential genetic algorithm SADE}

This method was proposed as an adaption of the differential evolution
after relatively long time of development. Its aim was to acquire a
method which is able to solve optimization problems on real domains
with a high number of variables (it was tested on problems with up to
$200$ variables). This algorithm combines the features of the
differential evolution with those of the traditional genetic
algorithms. It uses the simplified differential operator and the
algorithmic scheme similar to the standard genetic algorithm. \par

\subsection{The simplified differential operator} 
\begin{figure}[h]
\begin{center}
\scalebox{0.70}{
\includegraphics{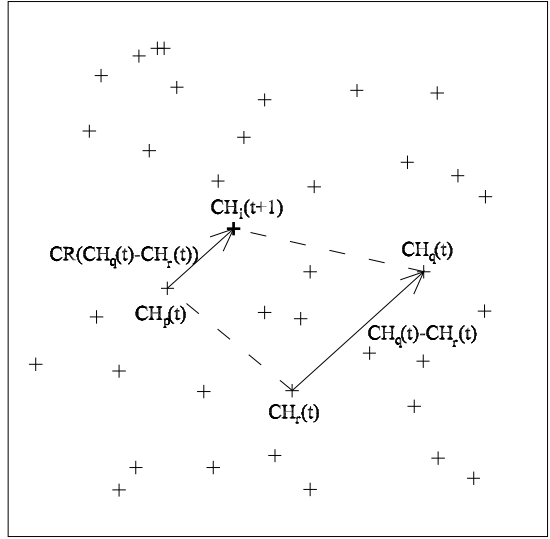}}
\end{center}
\caption{Geometric meaning of the simplified differential operator}
\label{operator}
\end{figure}

%
The simplified version of the differential operator taken from the
differential evolution is used for the same purpose as the
crossing-over in the standard genetic algorithm. Let (again) $CH_i(t)$
be the {\em i}-th chromozome in a generation~{\em t},
\begin{equation}
CH_i(t)=(ch_{i1}(t),ch_{i2}(t),\ldots,ch_{in}(t)), 
\end{equation}
where $n$ is the number of variables of the fitness function. Then the 
simplified differential operator can be written as
\begin{equation}
ch_{ij}(t+1)=ch_{pj}(t)+CR\left( ch_{qj}(t)-ch_{rj}(t) \right),
\end{equation}
where $ch_{pj}$, $ch_{qj}$ and $ch_{rj}$ are the {\em j}-th
coordinates of three randomly chosen chromozomes and $CR$ is so-called
{\em cross-rate\/}. Figure \ref{operator} shows the geometrical
meaning of this operator. Due to its independence on $j$, this
operator can be also rewritten in the vector form as
\begin{equation}
CH_i(t+1)=CH_p(t)+CR(CH_q(t)-CH_r(t)).    
\end{equation}

\subsection{The algorithmic scheme and the operators in detail}

Contrary to the differential evolution, the SADE method uses the 
algorithmic scheme very similar to the standard genetic algorithm:
\begin{enumerate}

\item As the first step, the initial population is generated randomly 
and the fitness function value is assigned to all chromozomes in the 
population.

\item Several new chromozomes are created using the mutation operators
- the mutation and the local mutation (their total number depends on
the value of a parameter called {\em radioactivity\/} -- it gives the
mutation probability).

\item Another new chromozomes are created using the simplified 
differential operator as was described above; the whole amount of 
chromozomes in the population is now doubled.

\item The fitness function values are assigned to all newly created 
chromozomes. 

\item The selection operator is applied to the double-sized
population. Hence, the amount of individuals is decreased to its
original value.

\item Steps 2-5 are repeated until a stopping criterion is reached.

\end{enumerate} 

Next, we describe the introduced operators in more detail:

\begin{description}
\item [mutation] -- if a certain chromozome $CH_i(t)$ was chosen to be 
mutated, a new random chromozome $RP$ is generated and the mutated one 
$CH_k(t+1)$ is computed using the following relation:
\begin{equation}
CH_k(t+1)=CH_i(t)+MR(RP-CH_i(t)),
\end{equation}
where $MR$ is a parameter called {\em mutation-rate\/},

\item [local mutation] -- if a certain chromozome was chosen to be 
locally mutated, all its coordinates are altered by a random 
value from a given (usually very small) range, 

\item [crossing-over] -- instead of traditional cross-over, the SADE
method uses the simplified differential operator described
above\footnote{Contrary to the binary genetic algorithm the real
encoded method may generate chromozomes outside the given domain. In
our implementation, this problem is solved by returning these
individuals to the feasible domain boundary.},

\item [selection] -- this method uses modified tournament strategy to
reduce the population size: two chromozomes are randomly chosen,
compared and the worse is rejected. Therefore, the population size is
decreased by one. This step is repeated until the population reaches
its original size\footnote{Contrary to the traditional tournament
strategy, this approach can ensures that the best chromozome will not
be lost even if it was not chosen to any tournament.}.

\end{description}

The detailed description of the SADE method including source codes in
C/C++ and the tests documentation for high-dimensional problems can be
obtained from the article \cite{Sade} and the web-page
\cite{Sade-WWW}. \par

\subsection{Testing and results}

The test computations were performed with the same functions and under
the same circumstances as in all previous cases. The population size
was set to $pop=10n$, which is the same value inherited from the
differential evolution. Another parameters were set, after several
trial runs, to $CR=0.2$ and $MR=0.5$, the local mutation range to
$0.25\%$ of the domain range of a corresponding variable and the {\em
radioactivity\/} was considered $20\%$. The results are shown in
Table~\ref{results}. \par

Similarly to the differential evolution method, the SADE algorithm
shows better behavior concerning the convergence rate and the
reliability than binary encoded methods. The overall reliability is
even better than for the differential evolution, but for the more
complicated problems, the number of the fitness calls is somewhat
bigger, even several times. This is caused by different behavior of
both methods from the character of the convergence process point of
view. While the differential evolution covers relatively large area of
the investigated domain during the whole process, the SADE algorithm
tends to create a cluster of individuals at a limited sub-area that
wanders through the domain. As a consequence, if the cluster is
deadlocked in a local extreme, it is necessary to wait until the
mutation gives a chance to escape to another sub-area with better
values. Of course, the probability of this effect is very low and
hence the algorithm must wait a long period of time. This effect
causes much worse results for problems with a rather large number of
local extremes. \par

\section{Improvement of the differential genetic algorithm to prevent 
the premature convergence - the CERAF method}

As already mentioned in the previous section, the SADE algorithm tends
to create clusters of chromozomes, which rather quickly wander through
the domain. This behavior somehow recalls gradient optimization
methods, however, with several differences: firstly, it operates with
more than one possible solution at a time, therefore it is able to
better locate the sub-area with the desired solution. Secondly, since
the changes of individuals are determined from their mutual distances,
this method is able to adapt the step size to reach an optimal
solution. \par

However, each time this method is caught in a local extreme, it has no
chance to escape unless a mutation randomly finds a sub-area with
better values. But the probability of this effect is very small,
especially for the high-dimensional problems. If the gradient
optimization methods are applied, this case is usually resolved by
so-called {\em multi-start} principle. It consists of restarting the
algorithm many times with different starting points. Similarly, any
type of a genetic algorithm could be restarted many
times. Nevertheless, the experience shows that there are functions
with so-called deceptive behavior, characterized by a high
probability that the restarted algorithm would fall again into the
same local extreme rather than focus on another sub-area. \par

Generally speaking, there are several solutions to this obstacle. All
of them are based on the leading idea of preventing the algorithm from
being trapped in the local extreme that has been already found and to
force the algorithm to avoid all of these. As the most natural way, we
tried some penalization that deteriorates the fitness function value
in the neighborhood of all discovered local extremes. However, this
approach did not approve itself -- if the shape of a penalization
function is not determined appropriately, new local extremes appear at
the boundary of a penalization function activity area. \par

As an alternative, the CERAF\footnote{Abbreviation of the French
expression {\em CEntre RAdioactiF} - the radioactivity center.} method
has been introduced. It produces areas of higher level of
``radioactivity'' in the neighborhood of all previously found local
extremes by increasing the mutation probability in these areas many
times (usually we set this probability directly to $100\%$). The range
of the radioactivity area (an {\em n}-dimensional ellipsoid) is set to
a certain percentage of the domain -- we denotes it as $RAD$.  The
time of stagnation that precedes the markup of a local extreme and
initiation of a radioactive zone is another parameter of the
method. Similarly to the living nature, the radioactivity in the CERAF
method is not constant in time but decreases in an appropriate way:
each time some individual is caught in that zone and mutated, the
radioactivity zone range is decreased by a small value\footnote{During
the numerical experiments it turned up that the chromozomes created by
the mutation parameter should not affect the radioactivity zone
range.} (for example $0.5\%$); this recalls the principle of
disintegration of a radioactive matter. The radioactive area never
disappears completely, so the chromozomes can never find the marked
local extreme again. \par

\subsection{The SADE algorithm extended by the CERAF method} 

Hereafter, the algorithmic scheme of the SADE method is supplied with
several steps of the CERAF method. It determines whether some
individuals got into any of the discovered ``radioactive zones'' and
if so, mutates them with a~high level of probability. Moreover, when
the algorithm stagnates too long, it declares a new radioactivity
area:

\begin{enumerate}

\item As the first step, the initial population is generated randomly 
and the fitness function value is assigned to all chromozomes in the 
population.

\item Several new chromozomes are created using the mutation operators
- the mutation and the local mutation (their total number depends on
the value of a parameter called {\em radioactivity\/} -- it gives the
mutation probability).

\item Another new chromozomes are created using the simplified 
differential operator as was described above; the whole amount of 
chromozomes in the population is now doubled.

\item If any radioactive zone already exists, each chromozome caught
in a radioactive area is, with a high probability, subjected to the
mutation operation.

\item Depending on the number of chromozomes determined in the
previous step, the ranges of radioactive zones are appropriately
decreased.

\item The fitness function values are assigned to all newly created 
chromozomes. 

\item The selection operator is applied to the double-sized
population. Hence, the amount of individuals is decreased to its
original value.

\item The number of stagnating generations is determined and if it
exceeds a given limit, the actual best solution is declared as the
center of the new radioactive area.

\item Steps 2-8 are repeated until a stopping criterion is reached.

\end{enumerate} 

Extensive test computations have shown that this methodology can be
considered as a universal technique capable of solving any multimodal
optimization problem provided that the method that is running
underneath (i.e. the algorithm that generates new chromozomes) has a
sufficient ability to find new possible solutions. In our case, the
SADE algorithm works as the ``exploration'' method. \par

\subsection{Test computations results}

For the purpose of the algorithm performance testing, the same
functions set was used. Also, all parameters of the SADE method rest
at the same values as before. The CERAF method parameters were
assigned the following values: $RAD $ is a $1/4$ of a domain range
(for each variable) and the mutation probability inside the
radioactive zones is considered $100\%$. The limit of stagnating
generations was set to $1700/pop$; this simple heuristic formula seems
to work well for a wide variety of problems. The results are given in
Table~\ref{results-CERAF} for the cases where the CERAF technology was
activated. In all the others the results are the same as for the
stand-alone SADE method. \par

\begin{table}[h]
\begin{center}
\begin{tabular}{lrrr}
\hline
\bf Test function & \bf Dimension & \bf Success rate \% & \bf Fitness 
calls \\
\hline
PShubert1 & 2 & 100 & 2388 \\
PShubert2 & 2 & 100 & 1014 \\
Shekel1 & 4 & 100 & 3942 \\
Shekel2 & 4 & 100 & 3746 \\
Shekel3 & 4 & 100 & 3042 \\
Hartman2 & 6 & 100 & 15396 \\
Brown1 & 20 & 100 & 137660 \\
F5n & 20 & 100 & 20332 \\
F10n & 20 & 100 & 200136 \\
F15n & 20 & 100 & 31574 \\
\hline
\end{tabular}
\end{center}
\caption{Results for the SADE/CERAF method.}
\label{results-CERAF}
\end{table}

Several interesting facts are evident when comparing these results 
with the previous cases:

\begin{itemize}

\item This method has reached the $100\%$ success for all test 
functions.

\item In many cases the number of fitness calls is the same as for the
single SADE algorithm; in those cases the CERAF technology was not
even activated because the simple algorithm found the global extreme
itself.

\item For the last (and the most complicated) functions {\em F5n},
{\em F10n} and {\em F15n}, the success has been improved from
$50-90\%$ to $100\%$, however, at the cost of slowing down the
computation. We consider indeed that the reliability of the method is
of greater value than the speed. These are the cases when the
algorithm extended by the CERAF method was able to continue searching
even after the previous simple method has been caught in a local
extreme hopelessly.

\item In several cases the computation was even accelerated by the
CERAF method, while the reliability was not decreased; in one
particular case (the {\em Hartman~2\/} function) the reliability was
even increased from $67\%$ to $100\%$. This may appear as a paradox,
because the CERAF method needs long periods of stagnation and repeated
optimum searching. The acceleration comes from the fact that the
method does not have to wait until the random mutation hits an area
with better values, but it is forced to start searching in a different
location.

\end{itemize}

\section{Conclusions}
%
As we have assumed, the presented results of test computations show
definitely that for the optimization of multimodal but still
continuous problems on real domains the evolutionary methods based on
real encoding and differential operators approve themselves much
better than traditional binary genetic algorithms, even when extended
by certain improvements. The real encoded algorithms produced better
results both in simple cases, where they have reached much better
(several times) convergence rates as well as in the complicated cases,
where the obtained results were very satisfactory from the reliability
point of view, even for functions where binary algorithms have
completely failed (e.g., the functions {\em Brown 1}, {\em Brown 3\/}
and {\em Hosc 45}). \par

\begin{table}[h]
\begin{center}
\begin{tabular}{lccccc}
\hline
\bf Function & \bf SGA & \bf EGA & \bf DE & \bf  SADE & \bf CERAF \\
\hline
F1                  & $\times$ & $\times$ & $\times$ & $\times$ & 
$\times$ \\
F3                  & $\times$ & $\times$ & $\times$ & $\times$ & 
$\times$ \\
Branin         &                   & $\times$ & $\times$ & $\times$ & 
$\times$ \\
Camelback & $\times$  & $\times$ & $\times$ & $\times$ & $\times$ \\
Goldprice    &                   & $\times$ & $\times$ & $\times$ & 
$\times$ \\
PShubert1   &                   & $\times$ &                   & 
$\times$ & $\times$ \\
PShubert2   &                   & $\times$ &                   & 
$\times$ & $\times$ \\
Quartic        &                   & $\times$ &                   & 
$\times$ & $\times$ \\
Shubert        &                   & $\times$ &                  & 
$\times$ & $\times$ \\
Hartman1   &                    & $\times$ & $\times$ & $\times$ & 
$\times$ \\
Shekel1        &                    & $\times$ &                   & 
$\times$ & $\times$ \\
Shekel2        &                    & $\times$ &                   & 
$\times$ & $\times$ \\
Shekel3        &                   & $\times$ &                    & 
$\times$ & $\times$ \\
Hartman2  &                 &                &                 &       
          & $\times$ \\
Hosc45 &                   &                    & $\times$ & $\times$ 
& $\times$ \\
Brown1 &                 &                    & $\times$ &             
&
$\times$ \\
Brown3 &                 &                    & $\times$ & $\times$ & 
$\times$ \\
F5n        &                 & $\times$ & $\times$ & $\times$ & 
$\times$ \\
F10n     &                  &                   &                  &   
                  & $\times$ \\
F15n     &                  & $\times$ & $\times$ & $\times$ & 
$\times$ \\
\hline
\end{tabular}
\end{center}
\caption{Comparison of reliability.}
\label{compare1}
\end{table}

The overall reliability-based comparison of all the tested methods is
provided in Table~\ref{compare1} ($\times$ marks the cases, where the
success was better than $95\%$). Note that the SADE algorithm extended
by the CERAF technology have achieved the $100\%$ success for all the
test functions. The next interesting result is that the single SADE
algorithm has approximately the same reliability as the binary
algorithm extended by several, rather sophisticated, improvements. The
reliability of a differential evolution is somehow fluctuating and the
standard binary algorithm does not show satisfactory behavior except
the most simple cases. \par

\begin{table}[ht]
\begin{center}
\vspace*{1cm}
\begin{tabular}{lccccc}
\hline
\bf Function & \bf SGA & \bf EGA & \bf DE & \bf  SADE & \bf CERAF \\
\hline
F1          &              &             & $\times$ &                  
 &                  \\
F3          &              &             &                   & 
$\times$ & $\times$ \\
Branin &              &             &                   & $\times$ &   
$\times$ \\
Camelback &      &             & $\times$ &                  &       
           \\
Goldprice    &     &              & $\times$ &                  &     
             \\
PShubert1   &     &              & $\times$ &                   &      
            \\
PShubert2   &     &              & $\times$ &                   &      
            \\
Quartic        &     &               & $\times$ &                   &  
                \\
Shubert        &     &               &                  & $\times$ & 
$\times$ \\
Hartman1   &    &                & $\times$ &                  &       
            \\
Shekel1        &     &               & $\times$ &                   &  
                 \\
Shekel2        &      &              & $\times$ &                   &  
                 \\
Shekel3        &      &              & $\times$ &                   &  
                 \\
Hartman2  &       &              &                 & $\times$ & 
 \\
Hosc45        &      &                & $\times$ &               &     
              \\
Brown1       &      &               & $\times$ &               
  &  \\
Brown3       &      &               & $\times$ &               &    
             \\
F5n              &      &               &                  & $\times$ 
&                 \\
F10n            &      &               & $\times$ &                   
&                \\
F15n            &      &               &                   & $\times$ 
&                 \\
\hline
\end{tabular}
\end{center}
\caption{Comparison of convergence rate.}
\label{compare2}
\end{table}

Table~\ref{compare2} shows the comparison of all methods from the
convergence rate point of view~($\times$ marks the case, where the
method reached the result at the shortest time). The differential
evolution seems to be the most effective (the fastest optimization
method). For other cases, the SADE method or its CERAF extended
version were the fastest ones. As could be seen, the binary algorithm
has never reached the best convergence rate with this test
computations. \par

\section*{Acknowledgement}
Authors would like to thank anonymous referees for their careful
revision and comments that helped us to substantially improve the
quality of the paper. This work has been supported by the Ministry of
Education, Youth and Sports of the Czech Republic (M{\v S}MT {\v C}R)
under project No.~210000003.

\appendix

\section{List of test functions}\label{sec:appa}

\begin{itemize}

\item F1: 
\begin{equation}
f(x)=2(x-0.75)^2+\sin (5\pi x-0.4\pi )-0.125 
\end{equation}
where 
\[
0\leq x\leq 1
\] 

\item F3: 
\begin{equation}
f(x)=-\sum_{j=1}^5 [j\sin [(j+1)x+j]]
\end{equation}
where
\[
-10\leq x\leq 10
\] 

\item Branin:
\begin{equation}
f(x,y)=a(y-bx^2+cx-d)^2+h(1-f)\cos x+h
\end{equation}
where
\[
a=1, b=5.1/4\pi ^2, c=5/\pi, d=6, 
\]
\[
h=10, f=1/8\pi, -5\leq x\leq 10, 0\leq y\leq 15
\]

\item Camelback:
\begin{equation}
f(x,y)=\left( 4-2.1x^2+\frac{x^4}{3}\right)x^2+xy+(-4+4y^2)y^2
\end{equation}
where
\[
-3\leq x\leq 3, -2\leq y \leq 2
\]

\item Goldprice:
\begin{eqnarray}
f(x,y)=& [ & 1+(x+y+1)^2(19-14x+3x^2-14y+6xy+3y^2)]\cdot \nonumber \\
& [ & 30+(2x-3y)^2(18-32x+12x^2+48y-36xy+27y^2)] \nonumber \\
\end{eqnarray}
where
\[
-2\leq x\leq 2, -2\leq y\leq 2
\]

\item PShubert 1 and 2:
\begin{eqnarray}
f(x,y)= & & \left\{ \sum_{i=1}^5 i\cos[(i+1)x+i]\right\}\cdot 
\nonumber \\
& & \left\{ \sum_{i=1}^5
i\cos[(i+1)y+i]\right\}+ \nonumber \\
& & \beta [(x-1.42513)^2+(y+0.80032)^2]
\end{eqnarray}
where
\[
-10\leq x\leq 10, -10\leq y\leq 10, 
\]
for PShubert1: \( \beta=0.5 \) \\
for PShubert2: \( \beta=1.0 \)

\item Quartic:
\begin{equation}
f(x,y)=\frac{x^4}{4}-\frac{x^2}{2}+\frac{x}{10}+\frac{y^2}{2}
\end{equation}
where
\[
-10\leq x\leq 10, -10\leq y\leq 10
\]

\item Shubert:
\begin{eqnarray}
f(x,y)=& & \left\{ \sum_{i=1}^5 i\cos[(i+1)x+i]\right\}\cdot \nonumber 
\\
& & \left\{ \sum_{i=1}^5 i\cos[(i+1)y+i]\right\}
\end{eqnarray}
where
\[
-10\leq x\leq 10, -10\leq y\leq 10
\]

\item Hartman 1:
\begin{equation}
f(x_1,x_2,x_3)=-\sum_{i=1}^4 c_i e^{-\sum_{j=1}^{3} 
a_{ij}(x_i-p_{ij})^2}
\end{equation}
where
\[
0\leq x_i\leq 1,  i=1,\ldots,3 
\]
\[
x=(x_1,\ldots,x_3), p_i=(p_{i1},\ldots,p_{i3}),
a_i=(a_{i1},\ldots,a_{i3})
\]
\begin{center}
\begin{tabular}{*8l}
\hline
$i$ & $a_{ij}$ & & & $c_i$ & $p_{ij}$ & & \\
\hline
1 & 3.0 & 10.0 & 30.0 & 1.0 & 0.36890 & 0.1170 & 0.2673 \\
2 & 0.1 & 10.0 & 35.0 & 1.2 & 0.46990 & 0.4387 & 0.7470 \\
3 & 3.0 & 10.0 & 30.0 & 3.0 & 0.10910 & 0.8732 & 0.5547 \\
4 & 0.1 & 10.0 & 35.0 & 3.2 & 0.03815 & 0.5743 & 0.8828 \\
\hline
\end{tabular}
\end{center}

\item Shekel 1,2 and 3:
\begin{equation}
f(x)=-\sum_{i=1}^{m} \frac{1}{(x-a_i)^T(x-a_i)+c_i}
\end{equation}
where
\[
0\leq x_j\leq 10, 
\]
for Shekel1: \( m=5 \), \\
for Shekel2: \( m=7 \), \\
for Shekel3: \( m=10 \) \\
\[
x=(x_1,x_2,x_3,x_4)^T, a_i=(a_{i1},a_{i2},a_{i3},a_{i4})^T
\]
\begin{center}
\begin{tabular}{*6l}
\hline
$i$ & $a_{ij}$ & & & & $c_i$ \\
\hline
1 & 4.0 & 4.0 & 4.0 & 4.0 & 0.1 \\
2 & 1.0 & 1.0 & 1.0 & 1.0 & 0.2 \\
3 & 8.0 & 8.0 & 8.0 & 8.0 & 0.2 \\
4 & 6.0 & 6.0 & 6.0 & 6.0 & 0.4 \\
5 & 3.0 & 7.0 & 3.0 & 7.0 & 0.4 \\
6 & 2.0 & 9.0 & 2.0 & 9.0 & 0.6 \\
7 & 5.0 & 5.0 & 3.0 & 3.0 & 0.6 \\
8 & 8.0 & 1.0 & 8.0 & 1.0 & 0.7 \\
9 & 6.0 & 2.0 & 6.0 & 2.0 & 0.5 \\
10 & 7.0 & 3.6 & 7.0 & 3.6 & 0.5 \\
\hline
\end{tabular}
\end{center}

\item Hartman 2:
\begin{equation}
f(x_1,\ldots,x_6)=-\sum_{i=1}^4 c_i e^{-\sum_{j=1}^6 a_{ij}(x_i-p_{ij})^2}
\end{equation}
where
\[
0\leq x_j\leq 1, j=1,\ldots,6
\]
\[
x=(x_1,\ldots,x_6), p_i=(p_{i1},\ldots,p_{i6}), a_i=(a_{i1},\ldots,a_{i6})
\]
\begin{center}
\begin{tabular}{*8l}
\hline
$i$ & $a_{ij}$ & & & & & & $c_{i}$ \\
\hline
1 & 10.00 & 3.00 & 17.00 & 3.50 & 1.70 & 8.00 & 1.0 \\
2 & 0.05 & 10.00 & 17.00 & 0.10 & 8.00 & 14.00 & 1.2 \\
3 & 3.00 & 3.50 & 1.70 & 10.00 & 17.00 & 8.00 & 3.0 \\
4 & 17.00 & 8.00 & 0.05 & 10.00 & 0.01 & 14.00 & 3.2 \\
\hline
\end{tabular}
\end{center}
\begin{center}
\begin{tabular}{*8l}
\hline
$i$ & $p_{ij}$ & & & & & \\
\hline
1 & 0.1312 & 0.1696 & 0.5569 & 0.0124 & 0.8283 & 0.5886 \\
2 & 0.2329 & 0.4135 & 0.8307 & 0.3736 & 0.1004 & 0.9991 \\
3 & 0.2348 & 0.1451 & 0.3522 & 0.2883 & 0.3047 & 0.6650 \\
4 & 0.4047 & 0.8828 & 0.8732 & 0.5743 & 0.1091 & 0.0381 \\
\hline
\end{tabular}
\end{center}

\item Hosc 45:
\begin{equation}
f(x)=2-\frac{1}{n!}\prod_{i=1}^n x_i
\end{equation}
where
\[
x=(x_1,\ldots,x_n), 0\leq x_i\leq i, n=10
\]

\item Brown 1:
\begin{eqnarray}
f(x)=& & \left[ \sum_{i\in J} (x_i-3)\right] ^2+ \nonumber \\
& & \sum_{i\in J} [10^{-3}(x_i-3)^2-(x_i-x_{i+1}) +e^{20(x_i-x_{i+1})}]
\end{eqnarray}
where
\[
J=\{ 1,3,\ldots,19 \},-1\leq x_i\leq 4, 1\leq i\leq 20, 
x=(x_1,\ldots,x_{20})^T
\]

\item Brown 3:
\begin{equation}
f(x)=\sum_{i=1}^{19} [(x_i^2)^{(x_{i+1}^2+1)}+(x_{i+1}^2)^{(x_i^2+1)}]
\end{equation}
\[
x=(x_1,\ldots,x_{20})^T, -1\leq x_i\leq 4, 1\leq i\leq 20
\]

\item F5n:
\begin{eqnarray}
& & f(x)= (\pi/20)\cdot \nonumber \\ 
& & \left\{ 10\sin^2(\pi y_1)+ \sum_{i=1}^{19}[(y_i-1)^2\cdot( 
1+10\sin^2(\pi y_i+1))]+(y_{20}-1)^2\right\}
\end{eqnarray}
where
\[
x=(x_1,\ldots,x_{20})^T,  -10\leq x_i\leq 10, y_i=1+0.25(x_i-1)
\]

\item F10n:
\begin{eqnarray}
& & f(x)=(\pi/20)\cdot \nonumber \\ 
& & \left\{ 10\sin^2(\pi x_1)+\sum_{i=1}^{19}
[(x_i-1)^2\cdot (1+10\sin^2(\pi
x_{i+1}))]+(x_{20}-1)^2\right\}
\end{eqnarray}
where
\[
x=(x_1,\ldots,x_{20})^T, -10\leq x_i\leq 10
\]

\item F15n:
\begin{eqnarray}
& & f(x)=(1/10)\cdot \nonumber \\ 
& & \left\{ \sin^2(3\pi x_1)+\sum_{i=1}^{19} [(x_i-1)^2(1+sin^2(3\pi
x_{i+1}))]+(1/10)(x_{20}-1)^2[1+\sin^2(2\pi x_{20})]\right\} \nonumber 
\\
\end{eqnarray}
where
\[
x=(x_1,\ldots,x_{20})^T, -10\leq x_i\leq 10
\]

\end{itemize}

\bibliographystyle{plain}
\bibliography{liter}

\end{document}